\documentclass[10pt,twocolumn,letterpaper]{article}

\usepackage[final]{cvpr}              
\usepackage{multirow}
\usepackage{upgreek}

\newcommand{\ourmethod}{$\text{AV}^2\text{A}$}

\definecolor{cvprblue}{rgb}{0.21,0.49,0.74}
\usepackage[pagebackref,breaklinks,colorlinks,allcolors=cvprblue]{hyperref}

\def\paperID{10543} 
\def\confName{CVPR}
\def\confYear{2025}

\title{Adapting to the Unknown: Training-Free Audio-Visual Event Perception with Dynamic Thresholds}

\author{
    Eitan Shaar$^{1\ast}$ \quad \quad
    Ariel Shaulov$^{2\ast}$ \quad \quad
    Gal Chechik$^{1,3}$ \quad \quad
    Lior Wolf$^2$ \\
    \\
    $^1$Bar-Ilan University \quad
    $^2$Tel-Aviv University \quad
    $^3$NVIDIA
}

\begin{document}
\maketitle

\def\thefootnote{*}\footnotetext{Equal Contribution. Order determined by a coin flip.}
\def\thefootnote{}\footnotetext{Corresponding authors: shaarei@biu.ac.il,arielshaulov@mail.tau.ac.il}

\begin{abstract}
In the domain of audio-visual event perception, which focuses on the temporal localization and classification of events across distinct modalities (audio and visual), existing approaches are constrained by the vocabulary available in their training data. This limitation significantly impedes their capacity to generalize to novel, unseen event categories. Furthermore, the annotation process for this task is labor-intensive, requiring extensive manual labeling across modalities and temporal segments, limiting the scalability of current methods. Current state-of-the-art models ignore the shifts in event distributions over time, reducing their ability to adjust to changing video dynamics. Additionally, previous methods rely on late fusion to combine audio and visual information. While straightforward, this approach results in a significant loss of multimodal interactions.  
To address these challenges, we propose \textbf{A}udio-\textbf{V}isual \textbf{A}daptive \textbf{V}ideo \textbf{A}nalysis (\ourmethod{}), a model-agnostic approach that requires no further training and integrates a score-level fusion technique to retain richer multimodal interactions. \ourmethod{} also includes a within-video label shift algorithm, leveraging input video data and predictions from prior frames to dynamically adjust event distributions for subsequent frames. Moreover, we present the first training-free, open-vocabulary baseline for audio-visual event perception, demonstrating that \ourmethod{} achieves substantial improvements over naive training-free baselines. We demonstrate the effectiveness of \ourmethod{} on both zero-shot and weakly-supervised state-of-the-art methods, achieving notable improvements in performance metrics over existing approaches.  Our code is available on \href{https://github.com/eitan159/AV2A}{Github}.
\end{abstract}    
\section{Introduction}
\begin{figure}[t]
  \centering
   \includegraphics[width=1.0\linewidth]{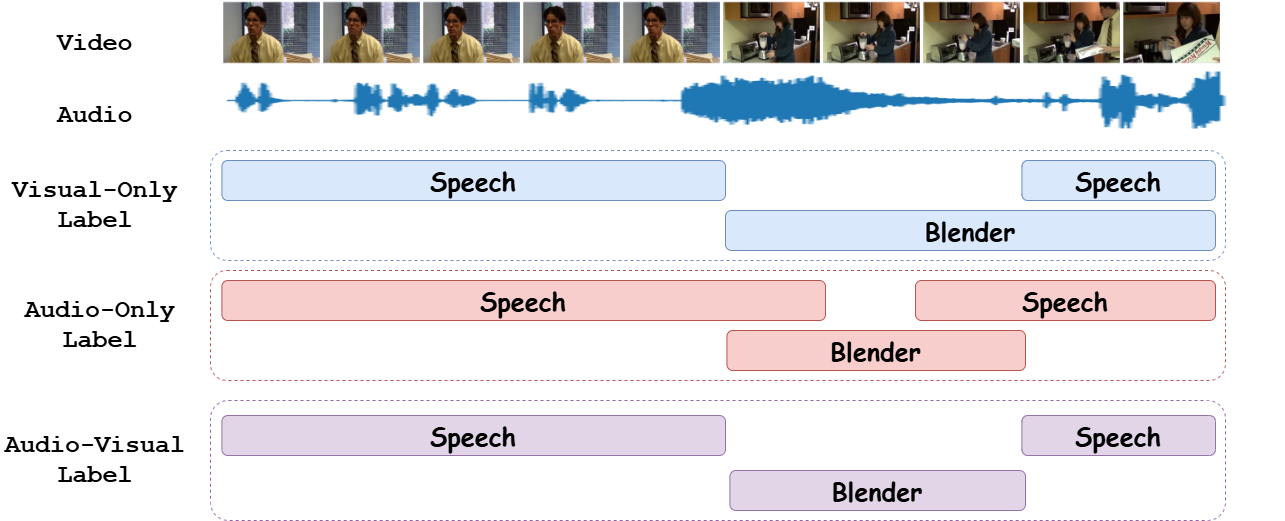}
  \caption{Overview of the AVVP task. Audio-visual event perception focuses on predicting the temporal boundaries of events within a video that are exclusively visible (shown in blue), exclusively audible (shown in red), or both audible and visible (shown in purple).}
   \label{fig:task_fig}
\end{figure}

\label{sec:intro}
Video event recognition, which involves the temporal localization and classification of events using both audio and visual cues (see Figure \ref{fig:task_fig}), is essential for a wide range of applications, including video content analysis, autonomous driving, multimedia indexing, and personalized content recommendation systems \cite{tian2018audio, wu2021exploring, gao2023collecting}. Despite the recognized importance of this task, current approaches face substantial limitations that hinder their effectiveness and adaptability in open-world scenarios.

A primary limitation in previous state-of-the-art methods \cite{gao2023collecting, cheng2022joint, jiang2022dhhn, rachavarapu2024weakly, lin2021exploring, mo2022multi, mo2022semantic} is their inability to dynamically adjust to evolving events in the video, leading to an oversight of changes in event distribution over time.
This issue stems from the reliance on fixed thresholds for event prediction, which restricts the model's capacity to adapt to evolving patterns and contextual shifts within the video sequence. Moreover, the current state-of-the-art methods often employ a late fusion strategy, where audio and visual information are processed independently and then merged by intersecting the predicted labels from both modalities to obtain audio-visual predictions.
While this approach simplifies model design, it limits the potential to capture the rich multimodal interactions needed for complex event detection \cite{gao2023collecting, mo2022multi}. 

Another major limitation of existing methodologies \cite{gao2023collecting, cheng2022joint, jiang2022dhhn, rachavarapu2024weakly, lin2021exploring, mo2022multi, mo2022semantic} is their reliance on fixed-vocabulary datasets. This reliance significantly restricts these models' ability to generalize to novel or evolving event categories—a capability that is essential for open-world environments where the range of possible event types is inherently diverse and continuously evolving \cite{wu2021exploring, cheng2022joint}. Addressing this challenge requires developing models that can move beyond pre-defined vocabularies and adapt to new events without extensive re-training. Lastly, 
the annotation process is labor-intensive, time-consuming, and costly, making it difficult to scale existing approaches to larger datasets or to adapt them for open-world deployment \cite{lin2021exploring}. 

To address these challenges, we propose a model-agnostic approach requiring no additional training, termed \textbf{A}udio-\textbf{V}isual \textbf{A}daptive \textbf{V}ideo \textbf{A}nalysis (\ourmethod{}). Our method integrates a score-level fusion mechanism that combines audio and visual cues at an initial processing stage, thereby preserving richer multimodal interactions essential for accurate and context-aware event recognition. This approach enhances the alignment of audio and visual information, improving the model’s capacity to recognize complex event patterns. Additionally, \ourmethod{} incorporates a novel within-video label-shift mechanism designed to operate within video contexts, leveraging both input video data and predictions from prior frames to dynamically adjust event distributions for subsequent frames, achieved via dynamic thresholds that adapt over time. 
By aligning model expectations with recurring or evolving events, this mechanism enables \ourmethod{} to capture temporal dependencies. In scenarios where events frequently recur or evolve, the dynamic thresholds capability ensures that the model remains adaptable and responsive to changing contexts. 
Furthermore, by leveraging foundation models \cite{zhu2023languagebind, elizalde2023clap, radford2021learning} we introduce the first training-free, open-vocabulary baseline
through \ourmethod{}. This approach circumvents the limitations of fixed vocabularies and reduces the need for additional data annotations. Our primary contributions are as follows: 
\begin{itemize} 
    \item We present \ourmethod{}, a model-agnostic approach that incorporates a score-level fusion strategy to combine audio and visual cues, preserving critical multimodal interactions. Additionally, our approach applies a within video label-shift mechanism that dynamically adjusts the predicted event distribution based on prior context, enhancing temporal adaptability. 
    \item We introduce the first training-free approach specifically tailored for open-world audio-visual event perception, capable of recognizing and adapting to unseen events without the constraints of a fixed vocabulary. 
    \item Through comprehensive evaluations on the AVE \cite{tian2018audio} and LLP \cite{tian2020unified} datasets, we demonstrate that our approach achieves state-of-the-art results in both zero-shot and training-free settings, showcasing its robustness and versatility. 
    \item We illustrate the efficacy of our label-shift mechanism through experimental applications on existing state-of-the-art weakly supervised methods, demonstrating consistent improvements across all metrics without additional training. These findings underscore the potential of leveraging temporal patterns directly within existing models.
\end{itemize}

\begin{figure*}[t]
  \centering
   \includegraphics[width=\textwidth]{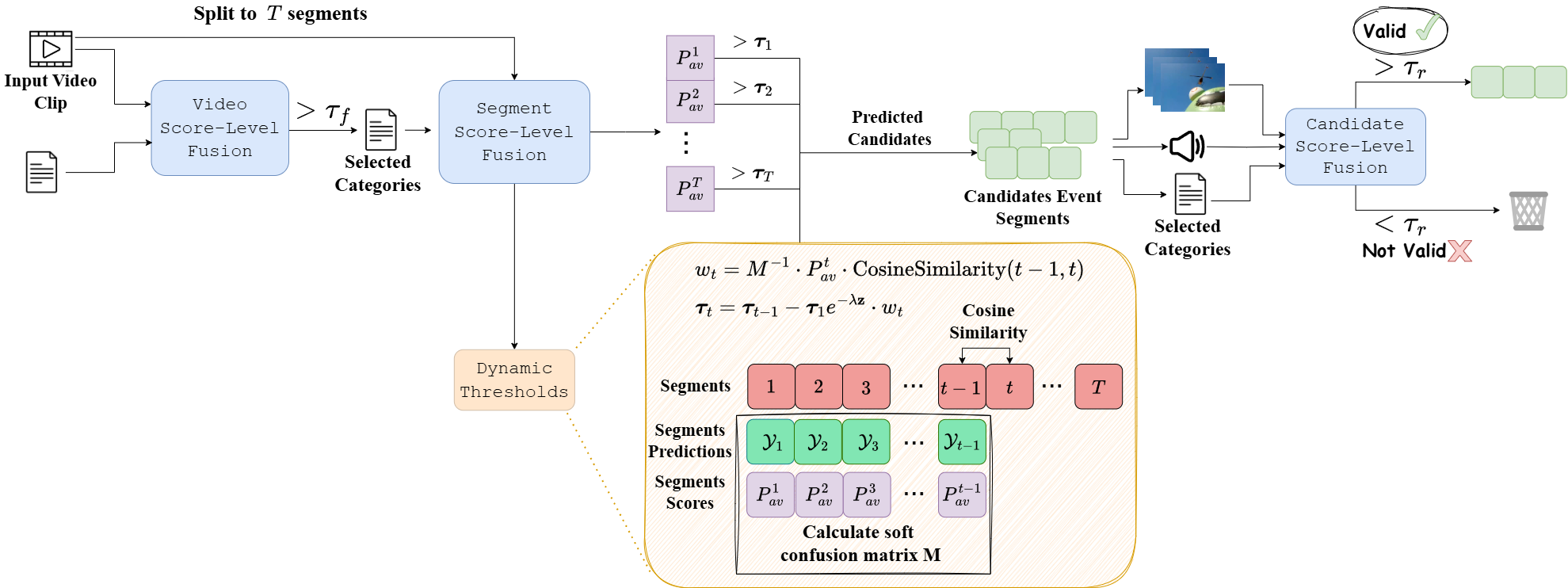}
   \caption{Overview of \ourmethod{}. The process begins with category selection, where the input video clip passes through a video-level score-level fusion module (blue) to select relevant categories based on a threshold $\tau_f$. These categories guide segment-level score-level fusion, where a dynamic threshold module (orange) updates thresholds $\boldsymbol{\tau}_t$ via our label-shift technique, using the soft confusion matrix $M$ from prior predictions $\mathcal{Y}_1, \dots, \mathcal{Y}_{t-1}$ and segment scores $P^1_{av}, \dots, P^{t-1}_{av}$, cosine similarity between segments and $P^t_{av}$. Finally, predicted candidates are validated against a confidence threshold $\tau_r$, retaining only those above it. This figure illustrates the process for audio-visual events; audio and visual events are handled similarly.}
   \label{fig:method}
\end{figure*}

\section{Related Work}
\paragraph{Weakly-supervised Audio-Visual Event Perception.}
To fully exploit the potential of both audio and visual modalities for weakly-supervised video analysis, \cite{tian2018audio} introduce the audio-visual event (AVE) localization task to leverage both modalities for weakly-supervised video understanding. In AVE, the model detects when an event is both audible and visible, pinpointing its temporal boundaries. Approaches often use attention strategies \cite{lin2020audiovisual, wu2022span, wu2019dual, xuan2020cross} or cross-modal interactions \cite{mahmud2023ave, ramaswamy2020makes, ramaswamy2020see, rao2022dual}, while others incorporate regularization like background suppression \cite{xia2022cross} and positive sample propagation \cite{zhou2021positive}.
Unlike AVE, the audio-visual video parsing (AVVP) task \cite{tian2020unified} accounts for misalignment between audio and visual signals. To address this, multi-modal multiple instance learning (MMIL \cite{tian2020unified}) frameworks have been employed \cite{gao2023collecting, cheng2022joint, jiang2022dhhn, rachavarapu2024weakly, lin2021exploring, mo2022multi, mo2022semantic}, enabling the model to identify relevant instances across modalities without the need for strict temporal synchronization. These methods typically include techniques such as audio-visual track swapping \cite{wu2021exploring} and cross-modality enhancement \cite{lin2021exploring}. 
Currently, most methods target either AVE or AVVP separately due to the different properties between the two tasks. 
Furthermore, all of the above methods are restricted to a closed vocabulary, requiring extensive training to recognize events, which limits their ability to generalize to new, unseen events.

\paragraph{Audio \& Visual Zero-Shot Methods.}
Audio-visual zero-shot learning was first introduced in \cite{parida2020coordinated} with the Coordinated Joint Multimodal Embedding (CJME) model, which aligns video, audio, and text in a shared feature space using triplet loss. Building on this, \cite{mazumder2021avgzslnet} developed the Audio-Visual Generalized Zero-shot Learning Network (AVGZSLNet), which employs a cross-modal decoder and composite triplet loss to enhance the alignment of audio and video embeddings with text. The Audio-Visual Cross-Attention (AVCA) framework \cite{mercea2022audio} further facilitates information exchange between modalities, achieving state-of-the-art performance in audio-visual zero-shot classification and introducing three novel benchmarks for AV-ZSL. Additionally, the Temporal Cross-Attention Framework (TCAF) \cite{mercea2022temporal} extends AVCA by leveraging temporal information. While \cite{duan2024cross} shows promising signs for zero-shot audio-visual event perception, the field remains largely underexplored, with significant opportunities for deeper exploration and advancement.

\paragraph{Label Shift Estimation.}
Label shift estimation refers to the task of inferring the label distribution in a target domain using labeled source data and unlabeled samples from the target domain \cite{garg2020unified}. Earlier methods   \cite{chan2005word, storkey2008training, zhang2013domain} often rely on explicit modeling of the conditional probability $P(X = x \mid Y = y)$, a requirement that becomes impractical for high-dimensional data, such as images.
More recently ~\cite{lipton2018detecting} proposed BBSE, a method that leverages a black box predictor $\hat{f}$ to estimate label shift between source and target distributions. Using predictions from $\hat{f}$, BBSE constructs a confusion matrix $M = p_s(\hat{y}, y) \in \mathbb{R}^{k \times k}$ on source data with $k$ classes,  where $y$ is the true label and $\hat{y}$ is the predicted label, which may be a hard label ($\hat{y} = \arg \max \hat{f}(x)$) or a soft probability vector, $\hat{f}(x) = [p(\hat{y} = 1|x), \dots, p(\hat{y} = k|x)]$, which is the distribution over classes predicted by $\hat{f}$ for a given input $x$.
The label shift estimate $\hat{w}$ is then calculated as $\hat{w} = M^{-1}\hat{\mu}$, where $M$ is the estimated confusion matrix and $\hat{\mu}$ represents the distribution of predicted labels in the target data, derived by applying $\hat{f}$ to the target samples. This correction factor $\hat{w}$ aligns the target label distribution with the source distribution. In contrast to prior approaches, we explore the integration of input data as a component within our label shift estimation method. To the best of our knowledge, this work is the first to adapt label shift estimation within video contexts.

\section{Method}

We present a framework for audio-visual event perception that is entirely training-free. Central to our approach is a dynamic thresholds mechanism, which adjusts threshold values over time-based on the input video data and previous segment predictions, see Figure ~\ref{fig:method}. The method enables precise detection of audio, visual, and audio-visual events. 

\paragraph{Problem Setup And Notations.}
Given an input video divided into $T$ non-overlapping segments, our objective is to temporally localize and identify audio, visual, and audio-visual events present for each segment $1 \leq t \leq T$. Each segment $t$ can contain zero or more events, which makes this a multi-label problem. The events to be identified are drawn from a set of 
$C$ possible categories, defined for each benchmark. The Given video consists of audio $A$ and video frames $V$: 
\
\[
A = \{a_1, a_2 \dots, a_T\}, \quad V = \{v_1, v_2, \dots, v_T\}
\]
\[
a_t \in \mathbb{R}^{1 \times F}, \quad v_t \in \mathbb{R}^{3\times H \times W}
\]
where $a_t$ and $v_t$ are the corresponding audio and video visual frame at segment $t$, $H$ and $W$ are the height and width for each frame, and $F$ is number of frequency bins for each audio snippet $a_t$.

We define two types of vector scores: video-level vector scores and segment-level vector scores. Both types are calculated by a foundation model (FM), combined with a sigmoid function $\sigma$ to address the multi-label classification problem. 
Our method operates in an open-vocabulary setting, where the label set $C$ is defined per video at inference time, even ones not seen during hyper-parameter tuning, unlike the baselines which are restricted to labels from a predefined set.
Instead, $C$ consists of user-defined textual descriptions, such as "an image of a military helicopter" or "the sound of helicopter rotor blades". 

Video-level scores are represented by $P_a$ and $P_v$ for audio and visual data, from the entire video’s audio $A$ and visual $V$. 
\begin{equation}
    \label{global_fm_scores}
    \centering
        P_a = \sigma(FM(C, A)), \quad P_v = \sigma(FM(C, V))
\end{equation}
\[
P_a, P_v \in [0,1]^{|C|}
\]
Segment-level vector scores capture more localized information. These are denoted as $P^t_a$ and $P^t_v$ for audio and visual scores, respectively, in each segment $t$ based on segment-specific audio $a_t$ and video frame $v_t$. 
\begin{equation}
    \label{segment_fm_scores}
    \centering
        P^t_a = \sigma(FM(C, a_t)), \quad P^t_v = \sigma(FM(C, v_t))
\end{equation}
\[
P^t_a, P^t_v \in [0,1]^{|C|}
\]
Each element in both types of score vectors represents the score that a specific category $c$ occurs. 

Effective audio-visual event recognition is based on the ability to leverage complementary information from both audio and visual modalities. Our score-level fusion is straightforward, we integrate audio and video scores using a weighted sum to create an audio-visual representation.
\begin{equation}
    \label{global_audio_visual}
    \centering
    P_{av} = \alpha P_a + (1 - \alpha) P_v
\end{equation}
\begin{equation}
    \label{segment_audio_visual}
    \centering
    P^t_{av} = \alpha P^t_a + (1 - \alpha) P^t_v
\end{equation}
\[
P_{av}, P^t_{av} \in [0,1]^{|C|}
\]
where $P_{av}, P^t_{av}$ are the audio-visual vector scores at the video-level and at segment-level $t$ respectively, and $\alpha\in [0,1]$ serves as a scalar weight that determines the contribution of the audio score $P_a, P^t_a$ relative to the visual score $P_v, P^t_v$. 

For generality, we will focus on identifying and localizing audio-visual events from this point forward. It is important to note that the processes for localizing and identifying audio and visual events are conducted in the same manner, using 
$\alpha = 1$ for audio events and $\alpha = 0$ for visual events in Equations \ref{global_audio_visual} and \ref{segment_audio_visual}.

In our framework, we first want to ensure that only the relevant event categories are retained.
We apply a selection step for the video-level vector $P_{av}$, discarding categories that do not exceed a predefined threshold $\tau_f$. 
This ensures us that only the most relevant event categories $\mathcal{\hat{C}}_{av}$ are preserved for further analysis.
\begin{equation}
    \label{filter_classes_eq}
    \centering
    \mathcal{\hat{C}}_{av} = \{ c \mid P_{av}[c] > \tau_f, \quad \forall c \in C\}\,,
\end{equation}
where $P_{av}[c]$ represents the audio-visual classification score for category 
$c$ is being presented in the video. 

To adapt to the content of the video domain, we apply thresholds for each category $c \in \mathcal{\hat{C}}_{av}$ that change over time. We initialize the threshold for each category with the same predefined value, denoted as $\tau^0$, which serves as the initial threshold at segment $t = 1$.
\[
\boldsymbol{\tau}_{t=1} = [\tau^0 \mid \forall c \in \mathcal{\hat{C}}_{av}]
\]
Then, we predict events for segment $t$ in the following way:
\begin{equation}
    \label{Label_Shift_Across_Time_eq}
    \mathcal{Y}_t =
    \begin{bmatrix}
        y^t_c \mid \forall c \in \mathcal{\hat{C}}_{av}
    \end{bmatrix}
\end{equation}
\[
\text{Where} \quad
y^t_c = 
\begin{cases}
    1 & \text{if } P^t_{av}[c] > \boldsymbol{\tau}_{t}[c], \\
    0 & \text{otherwise}
\end{cases}
\]
where $P^t_{av}[c]$ represents the audio-visual score
of category $c$ being presented at segment $t$ and $\boldsymbol{\tau}_{t}[c]$ is the category threshold for category $c$ at segment $t$. 

\begin{table}[t]
  \centering
  \resizebox{0.47\textwidth}{!}{%
  \begin{tabular}{@{}lccccc@{}}
    \toprule
    Method & $\alpha$ & $\tau^0$ & $\tau_r$ & $\tau_f$ & $\lambda$\\
    \midrule
    \midrule
        \ourmethod{} (CLIP \cite{radford2021learning} + CLAP \cite{elizalde2023clap}) & 0.45 & 0.75 & 0.75 & 0.5 & 1.0  \\
        \ourmethod{} (LanguageBind \cite{zhu2023languagebind}) & 0.5 & 0.75 & 0.75 & 0.55 & 2.5 \\
        \ourmethod{} (MGN-MA \cite{mo2022multi}) & 0.3 & 0.55 & - & - & 0.5 \\
        \ourmethod{} (JoMoLD \cite{cheng2022joint}) & 0.4 & 0.5 & - & - & 0.5 \\
        \ourmethod{} (CMPAE \cite{gao2023collecting}) & 0.45 & 0.5 & - & - & 0.5 \\
    \bottomrule
  \end{tabular}
  }
  \caption{Hyperparameters for each baseline using \ourmethod{}.}
\label{hyperparameters}
\end{table}

\begin{table*}[t]
    \centering
    \scalebox{0.90}{
    \setlength{\tabcolsep}{2pt}
    \begin{tabular}{@{}lcccccccccc@{}}
        \toprule
         & \multicolumn{2}{c}{Audio} & \multicolumn{2}{c}{Visual} & \multicolumn{2}{c}{Audio Visual} & \multicolumn{2}{c}{Type@AV} & \multicolumn{2}{c}{Event@AV}\\
        \cmidrule(lr){2-3} \cmidrule(lr){4-5} \cmidrule(lr){6-7} \cmidrule(lr){8-9} \cmidrule(lr){10-11}
        & segment & event & segment & event & segment & event & segment & event & segment & event \\
        \midrule
        \midrule
        CLIP \cite{radford2021learning} + CLAP \cite{elizalde2023clap} & 17.6 & 15.1 & 18.8 & 17.8 & 30.0 & 25.6 & 22.1 & 19.5 & 18.3 & 16.2 \\
        \quad with \ourmethod{}  & 32.7 $\pm$ .01 & 28.5 $\pm$ .01 & 43.5 $\pm$ .01 & 41.5 $\pm$ .01 & 55.1 $\pm$ .06 & 48.2 $\pm$ .04 & 43.8 $\pm$ .04 & 39.4 $\pm$ .01 & 34.7 $\pm$ .01 & 31.2 $\pm$ .01 \\
        \midrule
        LanguageBind \cite{zhu2023languagebind} & 20.3 & 17.5 & 21.3 & 20.0 & 24.3 & 21.0 & 22.0 & 19.5 & 20.8 & 18.4 \\ 
        \quad with \ourmethod{}  & \textbf{40.9} $\pm$ .09 & \textbf{35.9} $\pm$ .12 & \textbf{57.4} $\pm$ .04 & \textbf{54.4} $\pm$ .09 & \textbf{59.1} $\pm$ .1 & \textbf{52.3} $\pm$ .13 & \textbf{52.4} $\pm$ .04 & \textbf{47.5} $\pm$ .04 & \textbf{43.4} $\pm$ .04 & \textbf{38.9} $\pm$ .06 \\
        \bottomrule
    \end{tabular}
    }
    \caption{Comparison of training-free methods on the LLP dataset, reporting AVVP metrics.} 
\label{avvp_table}
\end{table*}

To modify our dynamic thresholds, we estimate for each segment $t$ the ratio $\hat{w}_t$ between the predictions generated from segments $1, \dots, t-1$ and those from segment $t$ across each category $c$.

To perform this estimation, 
we first define a soft confusion matrix $M$, where each entry represents the score that a given segment $t$ has the true category $c$ but it's predicted category is $\hat{c}$, averaged over all samples.
Let $M_p\in\mathbb{R}^{(t-1) \times \mathcal{|\hat{C}}_{av}|}$ be the matrix of prediction scores where each row is given by $P^t_{av}$ (Eq. \ref{segment_audio_visual}), which is the prediction score vector for segment 
$t$. We also define $M_y \in \mathbb{R}^{(t-1) \times \mathcal{|\hat{C}}_{av}|}$ as the matrix of the predicted categories where each row $\mathcal{Y}_t$ is predictions for segment $t$. 
The soft confusion matrix $M \in \mathbb{R}^{\mathcal{|\hat{C}}_{av}| \times \mathcal{|\hat{C}}_{av}|}$ is define as:
\[
M = \frac{1}{t-1} (M_p)^{T} \cdot M_y\,,
\]
i.e.,  each entry $M_{\hat{c}, c}$ in $M$ is given by:
\[
M_{\hat{c}, c} = \frac{1}{t-1} \sum^{t-1}_{i=1} P^i_{av}[\hat{c}] \cdot y^i_c \,.
\]
$M_{\hat{c}, c}$ captures the average score up to segment 
$t-1$ for instances where the model’s predicted score for the category 
$\hat{c}$ aligns with segments where the predicted category is 
$c$.
Additionally, we propose incorporating the input data as a scaling factor for $\hat{w}_t$. We use cosine similarity between the visual features of the current segment $t$, and the previous 
segment $t - 1$ as our scaling factor.
The estimated ratio $\hat{w}_t$ is computed as follows:
\begin{equation}
    \label{estimated_ratio}
    \centering
        \hat{w}_t = M^{-1}P^t_{av} \cdot \frac{x_t \cdot x_{t-1}}{\|x_t\| \|x_{t-1}\|} \quad,
\end{equation}
where $M$ is the estimated soft confusion matrix, $P^t_{av}$ is the score vector for segment $t$ and $x_t$ is denoted as the corresponding visual feature vector produced from the foundation model.
The estimated ratio $\hat{w}_t$, computed independently for each test video at inference time, quantifies the change in the score for each category $c$ between consecutive segments $t-1$ and $t$. Specifically, the ratio captures how the score vector $P^t_{av}$ for the current segment $t$ has shifted relative to the previous segments. The cosine similarity serves as a scaling factor, providing a measure of alignment between the feature vectors of adjacent segments. This ensures that if two segments are highly similar in content, any score changes are emphasized less, while more significant score changes in dissimilar segments are weighted more heavily. Thus, the ratio $\hat{w}_t$ not only reflects changes in category scores but also incorporates segment continuity, offering a refined measure of temporal dynamics in the event patterns across segments.

Following this, we update the category thresholds $\boldsymbol{\tau}_t$ for each segment $t$. This update employs an exponential decay function to gradually reduce the thresholds, ensuring that the adjustments become smaller over time and preventing excessive reductions as the process progresses. Here, $\boldsymbol{\tau}_1$ represents the initial threshold vector, while $\mathbf{z}$ is a vector tracking the cumulative count of occurrences for each category $c \in \mathcal{\hat{C}}_{av}$ up to time $t - 1$. The exponential decay, $e^{-\lambda \mathbf{z}}$, is applied element-wise, meaning that each component $z_i$ in $\mathbf{z}$ undergoes a separate transformation $e^{-\lambda z_i}$, allowing each category’s threshold adjustment to be scaled individually according to its occurrence frequency.
Here, $\lambda$ is the decay constant in the exponential decay function, controlling the rate at which threshold adjustments decrease.
\begin{equation}
    \label{exp_decay}
    \centering
    \boldsymbol{\tau}_t = \boldsymbol{\tau}_{t-1} - \boldsymbol{\tau}_1 e^{-\lambda \mathbf{z}} \cdot \hat{w}_t 
\end{equation}
\[
\boldsymbol{\tau}_t, \mathbf{z}, w_{t} \in \mathbb{R}^{1\times \mathcal{|\hat{C}}_{av}|}\quad,
\]

After obtaining all $\mathcal{Y}_1, \dots, \mathcal{Y}_T$, we define a candidate $S_k$, which is the predicted audio-visual localization for event category $c$, where 
$k$ represents the index of the multiple predicted candidates. $S_k$ is constructed to capture the maximal-length interval during which the score $P^t_{av}[c]$ remains above $\boldsymbol{\tau}_t[c]$ to each segment $t$, meaning that $y^t_c = 1$ holds true continuously throughout the entire candidate duration, such as: 
\[
S_k = (i_c^n, j_c^n, c) \quad \text{such that} \quad \exists n \leq T, \; \exists c \in \mathcal{\hat{C}}_{av}\quad,
\]
Where:
\[
i_c^n = \min \{ t \mid y^t_c = 1, \; t > j_c^{n-1} \}
\]
\[
j_c^n = \max \{ t \mid y^t_c = 1, \; t \geq i_c^n \}
\]
\[
\text{such that} \quad y^t_c = 1, \quad \forall t \in [i_c^n, j_c^n] \quad.
\]
Here, $i_c^n$ and $j_c^n$ denote the start and end indices of the candidate $S_k$, respectively, and $c$ represents the event category associated with the candidate $S_i$ and $n$ is the number of candidates for the same category $c$. \\ \\
To further refine the predicted candidates, each candidate $S_k$ is validated based on its associated audio $A'$ and video frames $V'$, where:
\[
A' = \{a_i, \ldots, a_j\}, \quad V' = \{v_i, \ldots, v_j\} \quad.
\]
Here, $i$ and $j$ represent the start and end timestamps of the predicted candidate. The candidate scores are computed as: 
 \begin{equation}
    \centering
    P'_a = \sigma(FM(\mathcal{\hat{C}}_{av}, A')), \quad P'_v = \sigma(FM(\mathcal{\hat{C}}_{av}, V'))\quad,
\end{equation}
where $P'_a$ and $P'_v$ denote the audio and visual scores for all categories $\mathcal{\hat{C}}_{av}$ between timestamps $i$, $j$. We compute the audio-visual score $P'_{av}$, which is computed as described in Equation~\ref{global_audio_visual} with $P'_a$ and $P'_v$. We apply a confidence threshold, $\tau_r$, to the audio-visual score $P'_{av}[c]$ for the predicted category $c$ associated with a candidate. If $P'_{av}[c]$ surpasses $\tau_r$, the candidate is deemed valid; otherwise, it is discarded. 
Our framework employs a total of $T \cdot |\hat{C}_{av}| + 2$ thresholds, specifically $ \{\boldsymbol{\tau}_1, \dots, \boldsymbol{\tau}_T\}, \tau_f,$ and $\tau_r$. The output of our framework consists of these valid candidates.

\begin{table*}[ht]
    \centering
    \setlength{\tabcolsep}{6pt} 
    \scalebox{0.93}{
    \begin{tabular}{@{}lcccccccccc@{}}
        \toprule
         & \multicolumn{2}{c}{Audio} & \multicolumn{2}{c}{Visual} & \multicolumn{2}{c}{Audio-Visual} & \multicolumn{2}{c}{Type@AV} & \multicolumn{2}{c}{Event@AV} \\
        \cmidrule(lr){2-3} \cmidrule(lr){4-5} \cmidrule(lr){6-7} \cmidrule(lr){8-9} \cmidrule(lr){10-11}
        & segment & event & segment & event & segment & event & segment & event & segment & event \\
        \midrule
        \midrule
        MGN-MA \cite{mo2022multi} & 60.7 & 51.0 & 55.5 & 52.4 & 50.6 & 44.4 & 55.6 & 49.3 & 57.2 & 49.2 \\
        \quad with \ourmethod{}'s dynamic thresholds & 60.8 & 51.0 & 55.6 & 52.7 & 51.9 & 45.5 & 56.1  & 49.7 & 57.3  & 49.3 \\
        \midrule
        JoMoLD \cite{cheng2022joint} & 61.3 & 53.9 & 63.8 & 59.9 & 57.2 & 49.6 & 60.8 & 54.5 & 59.9 & 52.5 \\
        \quad with \ourmethod{}'s dynamic thresholds & 61.4  & 53.9 & 64.1 & 61.0 & 57.7 & 50.9 & 61.1 & 55.2 & 60.0 & 52.7\\
        \midrule
        CMPAE \cite{gao2023collecting} & 64.4 & 57.5 & 66.1 & 63.3 & 59.5 & 52.4 & 63.3 & 57.7 & 62.9 & 56.3 \\
        \quad with \ourmethod{}'s dynamic thresholds & \textbf{64.7} & \textbf{58.0} & \textbf{66.3} & \textbf{63.4} & \textbf{59.9} & \textbf{52.7} & \textbf{63.6} & \textbf{58.0} & \textbf{63.0} & \textbf{56.7} \\
        \bottomrule
    \end{tabular}
    }
    \caption{Performance of state-of-the-art weakly supervised methods and their improvements with \ourmethod{}'s dynamic thresholds mechanism on the LLP dataset.}
    \label{baselines_table}
\end{table*}

\begin{table}
  \centering
  \begin{tabular}{@{}lc@{}}
    \toprule
    Method & Accuracy(\%) \\
    \midrule
    \midrule
        DG-SCT \cite{duan2024cross} & 64.7 \\ 
        \midrule
        CLIP \cite{radford2021learning} + CLAP \cite{elizalde2023clap} & 38.2 $\pm$ 2.87 \\ 
        \quad with \ourmethod{} & \underline{68.5} $\pm$ 0.16 \\
        \midrule
        LanguageBind \cite{zhu2023languagebind} & 32.9 $\pm$ 1.27 \\
        \quad with \ourmethod{} & \textbf{72.8} $\pm$ 0.14 \\
    \bottomrule
  \end{tabular}
  \caption{Comparison with zero-shot methods on AVE datasets, reporting AVEL metrics.}
\label{zero_shot_ave_table}
\end{table}
\section{Experiments}

We assess the performance of \ourmethod{} on two widely adopted datasets, AVE \cite{tian2018audio} and LLP \cite{tian2020unified}, through comprehensive experiments demonstrating its improvements in both training-free and zero-shot settings. Additionally, we apply our novel dynamic thresholds technique to state-of-the-art weakly supervised methods, achieving significant performance enhancements without additional training. An extensive ablation study further highlights the contribution of each component in our approach.

\subsection{Experimental setup }
\paragraph{Datasets.} We evaluate our approach on two datasets, AVE \cite{tian2018audio} and LLP \cite{tian2020unified}.

The Audio-Visual Event (AVE) dataset \cite{tian2018audio} consists of 4,143 YouTube videos spanning 28 categories derived from AudioSet \cite{gemmeke2017audio}, with each video clip lasting 10 seconds. The AVE \cite{tian2018audio} dataset focuses on recognizing audio-visual events that are both visible and audible across multiple time segments within a video, that is to say, it provides synchronized audio and visual signals, along with precise frame-level annotations that indicate the temporal locations of these events.

The Look, Listen, and Parse (LLP) dataset \cite{tian2020unified} is a comprehensive audio-visual benchmark designed for studying event perception in unconstrained, real-world videos. It comprises 11,849 YouTube video clips, each lasting 10 seconds and covering 25 diverse event categories that involve various audio-visual interactions. The dataset provides video-level event labels, including both single and multiple overlapping events. The LLP dataset suits weakly-supervised learning, providing video-level annotations without precise event boundaries.
\begin{figure*}[t]
   \includegraphics[width=\textwidth]{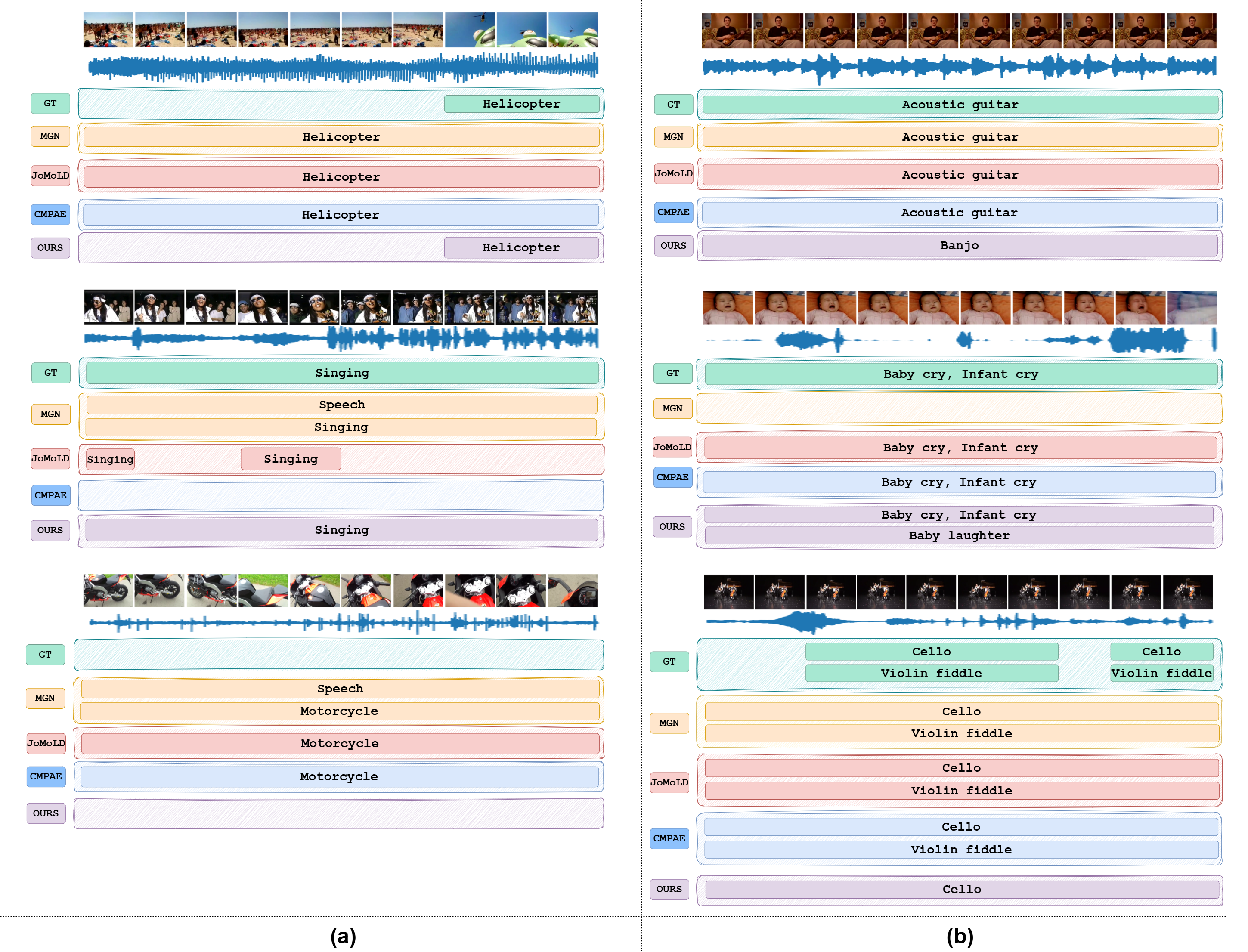}
   \caption{Performance analysis of \ourmethod{}, based on the LanguageBind \cite{zhu2023languagebind} model, showcasing predictions on audio-visual events, specifically those occurring simultaneously in both audio and video. Comparisons highlight (a) improvements and (b) failure-cases relative to state-of-the-art weakly supervised baselines.}
   \label{fig:wins_losses}
\end{figure*}

\paragraph{Evaluation Metrics.}\label{Evaluation_Metrics} 
Consistent with previous work, we evaluate using F1-scores for audio, visual, and audio-visual events at both segment and event levels. Additionally, we report aggregate metrics, Type@AV and Event@AV, computed at segment and event levels as well. For a comprehensive explanation of these metrics, refer to \cite{tian2020unified}. For the AVE dataset, in line with \cite{tian2018audio}, we use overall accuracy as the evaluation metric. 

For hyperparameter optimization, we performed a search over 200 samples from the LLP validation set for the training-free methods, while using the full validation set of LLP for the weakly-supervised methods. Since only \ourmethod{}'s dynamic threshold mechanism is applied, optimizing the $\tau_r$ and $\tau_f$ values is unnecessary for the MGN-MA \cite{mo2022multi}, JoMoLD \cite{cheng2022joint}, and CMPAE \cite{gao2023collecting} methods. All hyperparameters used for each baseline are listed in Table~\ref{hyperparameters}.

\subsection{Comparison with State-of-the-art Methods}
\paragraph{Baselines.}\label{Baselines} 
In our approach to training-free compression, we employ three foundational models to create two training-free baseline methods, upon which we implement \ourmethod{}. The first baseline employs the LanguageBind model \cite{zhu2023languagebind}, which provides both text-visual and text-audio similarity measurements. The second baseline combines the CLIP \cite{radford2021learning} and CLAP \cite{elizalde2023clap} models, where CLIP captures text-visual similarities, and CLAP supplies text-audio similarities; we refer to this baseline as CLIP-CLAP. For the evaluation of the CLIP-CLAP and LanguageBind baselines, a single static threshold is applied to predict categories across segments, creating two naive baselines. For weakly-supervised compression, we adopt three methodologies: JoMoLD \cite{cheng2022joint}, which performs joint modality learning for improved cross-modal representations; MGN-MA \cite{mo2022multi}, which leverages multi-granularity networks to enhance modality alignment; and the state-of-the-art method CMPAE \cite{gao2023collecting}, that integrates adaptive encoding to capture complex, high-quality cross-modal associations.

\begin{table*}[h!]
    \centering
    \setlength{\tabcolsep}{4pt}
    \begin{tabular}{@{}lcccccccccc@{}}
        \toprule
        Method & \multicolumn{2}{c}{Audio} & \multicolumn{2}{c}{Visual} & \multicolumn{2}{c}{Audio Visual} & \multicolumn{2}{c}{Type@AV} & \multicolumn{2}{c}{Event@AV}\\
        \cmidrule(lr){2-3} \cmidrule(lr){4-5} \cmidrule(lr){6-7} \cmidrule(lr){8-9} \cmidrule(lr){10-11}
        & segment & event & segment & event & segment & event & segment & event & segment & event \\
        \midrule
        \midrule
        \ourmethod{}(LanguageBind \cite{zhu2023languagebind}) & \textbf{40.9} & \textbf{35.9} & \textbf{57.4} & \textbf{54.4} & \textbf{59.1} & \textbf{52.3} & \textbf{52.4} & \textbf{47.5} & \textbf{43.4} & \textbf{38.9} \\
        \quad  w/o cosine similarity & 40.8 & \textbf{35.9} & 57.3 & 54.3 & 58.8 & 51.8 & 52.3 & 47.3 & 43.3 & 38.8 \\ 
        \quad  w/o dynamic thresholds & 40.7 & 35.1 & 56.9 & 55.2 & 55.1 & 47.2 & 52.2 & 47.2 & 43.0 & 37.8 \\
        \quad  w/o refine segments & 38.9 & 34.1 & 53.6 & 50.9 & 57.9 & 51.1 & 50.2 & 45.4 & 41.4 & 37.0 \\
        \quad w/o relevant class selection & 33.4 & 29.3 & 40.2 & 38.4 & 39.7 & 34.3 & 37.8 & 34.0 & 34.8 & 31.4 \\ 
         
        \bottomrule
    \end{tabular}
    \caption{Ablation study results on the LLP dataset.}
\label{ablation_table}
\end{table*}

\vspace{-1em}
\paragraph{AVVP Evaluation.} 
In this experiment, we employ two baselines, CLIP-CLAP and LanguageBind, which are constructed as described in Section \ref{Baselines}. As shown in Table \ref{avvp_table}, applying \ourmethod{} to the baselines CLIP-CLAP and LanguageBind results in metric improvements across the board. Specifically, when \ourmethod{} is integrated with the naive CLIP-CLAP baseline, it achieves a 21.7\% improvement in segment-level Type@AV, which measures average segment performance across audio, visual, and audio-visual settings. Additionally, \ourmethod{} enhances performance by 19.9\% in event-level Type@AV, which evaluates the average event performance across these setups. Furthermore, \ourmethod{} improves the naive LanguageBind baseline by 30.4\% and 28\% in segment-level and event-level Type@AV, respectively. 

We further validate \ourmethod{}'s dynamic thresholds mechanism on the AVVP benchmark by applying it on top of current state-of-the-art weakly supervised methods, specifically, CMPAE \cite{gao2023collecting}, JoMoLD \cite{cheng2022joint}, and MGN-MA \cite{mo2022multi}. In this experiment, we apply the dynamic thresholds mechanism (Eq. \ref{exp_decay}) to these methods, as they already employ a relevant class selection stage that closely aligns with our approach; however, these methods used static thresholds to predict events over time. The results, presented in Table \ref{baselines_table}, indicate that our dynamic thresholds outperform the static thresholds technique without additional training, establishing a new state-of-the-art in video parsing.
To implement our dynamic threshold mechanism in the audio-visual regime, it is essential to apply our score-level fusion stage (Eq.\ref{segment_audio_visual}), as the audio-visual scores required for the dynamic thresholds mechanism are unavailable in the methods standard form due to their late-fusion strategy. 

\vspace{-1em}
\paragraph{AVE Evaluation.}
As shown in Table ~\ref{zero_shot_ave_table}, we compare \ourmethod{} with three zero-shot baselines: the CLIP-CLAP baseline \cite{radford2021learning, elizalde2023clap}, the LanguageBind baseline \cite{zhu2023languagebind}, and the state-of-the-art DG-SCT method \cite{duan2024cross}. Notably, \ourmethod{} outperforms DG-SCT \cite{duan2024cross}, achieving improvements of 3.8\% and 8.1\% when built upon the CLIP-CLAP and LanguageBind baselines, respectively. 
To ensure a fair zero-shot comparison, we used the same hyperparameters identified during tuning on the LLP validation set, as detailed in Section \ref{Baselines}.
Please note that we did not apply \ourmethod{} on the DG-SCT baseline, as the trained checkpoints and features were not made available.

\vspace{-1em}
\paragraph{Method Efficiency.} In our experiments, inference times across all weakly-supervised methods were of the same order of magnitude, with our method achieving $\sim$ 2.49 seconds per video. However, our primary efficiency advantage lies in the fact that \ourmethod{} requires no additional training, unlike the weakly-supervised baselines. For comparison, the weakly-supervised methods trainable parameters are as follows: MGN-MA \cite{mo2022multi} has $\sim 4.3M$, JoMoLD \cite{cheng2022joint} has $\sim 4.5M$, and CMPAE \cite{gao2023collecting} has $\sim 5.6M$. In addition to the trainable parameters, the above methods employ three pre-trained, fixed models to generate input features: ResNet152 \cite{he2016deep}, VGGish \cite{hershey2017cnn}, and R(2+1)D \cite{tran2018closer}. This procedure adds $\sim$ 2.68 seconds to the inference time per video.

\subsection{Ablation study}
We conducted an ablation study using the LanguageBind baseline \cite{zhu2023languagebind} to investigate the contributions of each component of our method. 
The results, detailed in Table \ref{ablation_table}, illustrate the impact of these components on performance. The first row presents our complete method, with the LanguageBind model \cite{zhu2023languagebind}. The second row of the table represents the model configuration without cosine similarity in the label shift method, effectively demonstrating the method’s performance when excluding the cosine similarity between adjacent frames. The third row highlights the effect of omitting the dynamic threshold mechanism, employing fixed thresholds instead. Additionally, the fourth row reports the outcome when the refinement of segments is removed, resulting in candidate event segments being used as final predictions. Lastly, we examined the model without the relevant class selection process, using the entire class set from the outset. Note that in each row, only the specified stage is pruned, while the remainder of the pipeline remains unchanged. The findings indicate that each component of our method contributes positively to the overall performance, reinforcing the importance of their inclusion.
\section{Limitations}

Our reliance on foundation models for extracting audio-visual similarities may introduce biases and limit the accuracy of short-duration or rapid events (Figure \ref{fig:wins_losses}). These models, trained on longer, distinct events, struggle with subtle changes. Since \ourmethod{} requires no training or fine-tuning, improvements in foundation models should enhance audio-visual perception.

\section{Conclusion} 
In this paper, we introduce the first training-free, open-vocabulary approach for audio-visual perception via \ourmethod{}, a model-agnostic method for robust event perception. \ourmethod{} improves weakly-supervised baselines by replacing static thresholds with a within-video label shift mechanism, enhancing adaptability to temporal variations in event distribution. Applied to zero-shot and weakly-supervised settings, \ourmethod{} achieves new state-of-the-art results without additional training.

\section{Acknowledgments}
This work was supported by the Tel Aviv University Center for AI and Data Science (TAD), and by a Vatat data-science grant to GC and the Bar-Ilan University data-science institute. 

\vspace{-1em}
{
    \small
    \bibliographystyle{ieeenat_fullname}
    \bibliography{main}
}

\clearpage
 \appendix
\twocolumn[{%
    \begin{center}
        {\LARGE \textbf{Supplementary Material: \\ Adapting to the Unknown: Training-Free Audio-Visual Event Perception with Dynamic Thresholds}} \\[10pt]
        \vspace{5pt}
    \end{center}
}]

\section{Score-Level Fusion}
We conducted an experiment to evaluate the impact of a straightforward score-level fusion approach on existing baselines. The score-level fusion technique involves integrating audio and visual vectors through a weighted sum, controlled by a parameter $\alpha$. To ensure minimal modifications, we exclusively searched for the optimal $\alpha$ value through hyperparameter tuning on the LLP validation set, without retraining the models or adjusting any other parameters. As shown in Table \ref{early_fusion}, this simple score-level fusion (Eq. \ref{segment_audio_visual}) method improved the performance of audio-visual results, demonstrating its effectiveness in enhancing baseline models without additional computational overhead.
Notably, this analysis focuses exclusively on audio-visual results, as the score-level fusion technique does not alter the outcomes of individual audio or visual modalities.

\begin{table}[h!]
    \centering
    \setlength{\tabcolsep}{4pt}
    \begin{tabular}{@{}lcccc@{}}
        \toprule
        Method & \multicolumn{2}{c}{Audio Visual}\\
        \cmidrule(lr){2-3}
        & segment & event \\
        \midrule
        \midrule
        MGN-MA  \cite{mo2022multi} & 50.6 & 44.4 \\
        \quad + score-level fusion & 51.8 & 45.7 \\
        \midrule
        JoMoLD \cite{cheng2022joint} & 57.2 & 49.6 \\
        \quad + score-level fusion & 57.7 & 51.0 \\
        \midrule
        CMPAE \cite{gao2023collecting} & 59.5 & 52.4 \\
        \quad + score-level fusion & \textbf{59.8} & \textbf{52.6} \\
        \midrule
        LanguageBind with \ourmethod{} & 57.1 & 50.6 \\
        \quad + score-level fusion & 59.1 & 52.3 \\
        \midrule
        CLIP+CLAP with \ourmethod{} & 51.8 & 45.7 \\
        \quad + score-level fusion & 55.1 & 48.2 \\
        \bottomrule
    \end{tabular}
    \caption{The effect of score-level fusion instead of late fusion for the current state-of-the-art weakly supervised methods and for our training-free method.}
\label{early_fusion}
\end{table}

\section{Comparison to Weakly-Supervised Methods}
We compare our training-free method performance against weakly supervised baselines across segment-based and event-level metrics (see Table \ref{baselines_table_supp}), analyzing its strengths in multimodal fusion despite the absence of training. The results are reported for 1124 out of 1200 test videos, as only these videos are accessible on the internet. For the audio-visual segment-based score, our training-free method is on par with CMPAE \cite{gao2023collecting} and better than JoMoLD \cite{cheng2022joint} and MGN-MA \cite{mo2022multi}. For the event score, it is slightly better than CMPAE and much better than the rest. In separate audio and visual metrics, our method is inferior to the weakly supervised methods, emphasizing its multimodal nature.

\section{Linear Classification with CLIP/CLAP}
To further analyze the effectiveness of the features extracted from CLIP and CLAP, we trained a linear classifier to predict the category of events per second. This evaluation serves as a measure of the linear separability of these features and provides information on their suitability for event classification in a weakly supervised setting.
\paragraph{Experimental Setup.} We trained a fully supervised linear classifier using the features from CLIP and CLAP. The classifier was optimized to predict the event category for each second of the video, thereby assessing the discriminative power of these features at a fine temporal granularity. The LLP dataset was used for evaluation, and we report both segment-level and event-level performance. 

The results shown in Table \ref{Linear_classification} indicate that CLIP features exhibit stronger linear separability than CLAP features for event classification in this setting. This suggests that vision-language models such as CLIP encode more discriminative information that can be leveraged for event recognition with a simple linear probe. The lower performance of CLAP features may be attributed to the nature of audio embeddings, which might require more complex modeling techniques beyond linear classification to effectively capture event distinctions.

\begin{table}[h!]
    \centering
    \setlength{\tabcolsep}{4pt}
    \begin{tabular}{@{}lcccc@{}}
        \toprule
        Method & segment & event \\
        \midrule
        \midrule
        CLIP \cite{radford2021learning} & 33.9 & 32.1 \\
        \midrule
        CLAP \cite{elizalde2023clap} & 27.2 & 25.9 \\
        \bottomrule
    \end{tabular}
    \caption{Linear classifier performance on CLIP and CLAP features for event classification on the LLP dataset.}
\label{Linear_classification}
\end{table}

\begin{table*}[t!]
    \centering
    \setlength{\tabcolsep}{4pt} 
    \scalebox{0.93}{
    \begin{tabular}{@{}lcccccccccc@{}}
        \toprule
         & \multicolumn{2}{c}{Audio} & \multicolumn{2}{c}{Visual} & \multicolumn{2}{c}{Audio-Visual} & \multicolumn{2}{c}{Type@AV} & \multicolumn{2}{c}{Event@AV} \\
        \cmidrule(lr){2-3} \cmidrule(lr){4-5} \cmidrule(lr){6-7} \cmidrule(lr){8-9} \cmidrule(lr){10-11}
        & segment & event & segment & event & segment & event & segment & event & segment & event \\
        \midrule
        \multicolumn{11}{c}{\textbf{Weakly Supervised Methods}} \\
        \midrule
        MGN-MA \cite{mo2022multi} & 60.3 & 50.5 & 55.3 & 52.2 & 50.1 & 44.0 & 55.3 & 48.9 & 56.9 & 48.8 \\
        \midrule
        JoMoLD \cite{cheng2022joint} & 61.1 & 53.7 & 63.5 & 59.8 & 56.8 & 49.3 & 60.5 & 54.2 & 59.7 & 52.3 \\
        \midrule
        CMPAE \cite{gao2023collecting} & \textbf{64.1} & \textbf{57.2} & \textbf{66.1} & \textbf{63.3} & \textbf{59.1} & 52.2 & \textbf{63.3} & \textbf{57.7} & \textbf{62.9} & \textbf{56.3} \\
        \midrule
        \multicolumn{11}{c}{\textbf{Training-Free Methods}} \\
        \midrule
        LanguageBind+\ourmethod{} & 40.9 {\scriptsize $\pm$ .09} & 35.9 {\scriptsize $\pm$ .12} & 57.4 {\scriptsize $\pm$ .04} & 54.4 {\scriptsize $\pm$ .09} & \textbf{59.1} {\scriptsize $\pm$ .1} & \textbf{52.3} {\scriptsize $\pm$ .13} & 52.4 {\scriptsize $\pm$ .04} & 47.5 {\scriptsize $\pm$ .04} & 43.4 {\scriptsize $\pm$ .04} & 38.9 {\scriptsize $\pm$ .06} \\
        \bottomrule
    \end{tabular}
    }
    \caption{Performance of state-of-the-art weakly supervised methods in comparison to our training-free method on the LLP test set.}
    \label{baselines_table_supp}
\end{table*}

\end{document}